# Efficient algorithm for estimation of qualitative expected utility in possibilistic case-based reasoning

Jakub Brzostowski and Ryszard Kowalczyk
Faculty of Information and Communication Technologies
Swinburne University of Technology
Hawthorn, Victoria 3122, Australia

## Abstract

We propose an efficient algorithm for estimation of possibility based qualitative expected utility. It is useful for decision making mechanisms where each possible decision is assigned a multi-attribute possibility distribution. The computational complexity of ordinary methods calculating the expected utility based on discretization is growing exponentially with the number of attributes, and may become infeasible with a high number of these attributes. We present series of theorems and lemmas proving the correctness of our algorithm that exibits a linear computational complexity. Our algorithm has been applied in the context of selecting the most prospective partners in multi-party multi-attribute negotiation, and can also be used in making decisions about potential offers during the negotiation as other similar problems.

## 1 Introduction

The classical decision theory employs the notion of probabilistic expected utility [14]. In that model the uncertainty about the actual state is modelled by a probability distribution $p : D \to [0,1]$ that assigns probability values to each possible state. It is also assumed that the utility function that encodes the decision maker's preferences is given by $u : D \to R$. A probability distribution $p_d$ is assigned to each decision $d$. The decision that maximizes the expected utility can be chosen as follows:

$$EU(d) = \sum_{x \in D} p_d(x) u(x)$$

This approach requires the decision maker's knowledge to be represented by probability distributions. However, given information may be not sufficient for deriving an appropriate probability distribution in particular in multi-dimensional decision problems [13]. Therefore some non-probabilistic models of uncertainty such as Shafer's theory of evidence, possibility theory and nonmonotonic logic were introduced [9]. Possibility theory is very suitable as a basis for qualitative decision theory [7]. In that theory the notion of possibility distribution is the counterpart of probability distribution in probability theory. The possibility distribution function assigns to each possible outcome a level of plausibility: $\pi : D \to [0,1]$. As in classical approaches a possibility distribution may be assigned to each of decision maker's actions (decisions), and the task is to make the best decision. The utility function is also defined $u : D \to [0,1]$. The optimal decision can be found by maximising a qualitative utility function that can be defined according to the pessimistic or optimistic criteria [7] as follows:

$$QU^-(\pi|u) = min_{x \in D} max(n(\pi(x)), u(x))$$

$$QU^+(\pi|u) = max_{x \in D} min(\pi(x), u(x))$$

The possibility based decision theory may be useful in situations where we do not have numerical utilities about consequences of actions [12]. Sometimes the possibility distribution may be derived by case-based reasoning [8] however most approaches consider only one-dimensional space. In [11] [10] authors apply it in auctions to determine the succesful single-attribute bids. The possibilistic case-based reasonng is further developed by Dubois et al. [5] The possibilistic multi-attribute decision theory has its roots in fuzzy maxmin optimization paradigm initiated by Bellman and Zadeh [1]. The fuzzy approach to multiple critieria decision-making combined together with the constraint-directed methodology for combinatorial problem solving has also been studied in the context of flexible (or fuzzy) constraint satisfaction problems (FCSP) [4].

We propose an algorithm that is appropriate for sit-uations where the distribution has been constructed by case-based reasoning in a multi-dimensional space and the optimistic criterion is used to calculate the expected utility. The possibility based reasoning from historical data allows deriving the possibility distribution in situations where there are not many historical cases. The probabilistic approach, in opposite, requires a vast history of previous interactions that may not be available in some situations. The algorithm may be very useful as a part of decision making mechanisms of negotiating agents. In order to be efficient the agents in Multi-Agent Systems need to make the decisions fast. Very often the agent has to construct many possibility distributions and derive the expected utility for each of them. In such situations its desirable to design an algorithm that could estimate the expected utility in an efficient way. We propose an efficient algorithm performing this task without even constructing the distribution functions.

The next section presents a potential application of the algorithm for expected utility estimation. Section 3 presents the common definition of similarity relations used for case-based reasoning and the rule of this reasoning. In section 4 we explore the shapes of $\alpha$-cuts of the functions obtained by case-based reasoning and present appropriate theorems. In section 5 the Pareto order and monotonicity of multi-dimensional functions are defined. Section 6 describes the transformation of the density function into the corresponding distribution. Section 7 explores the shapes of $\alpha$-cuts of the transformed function and the Pareto frontier of these $\alpha$-cuts. Section 8 presents a theorem that is a basis for the algorithm calculating an estimation of the expected utility summarised and evaluated in Section 9. Finally, concluding remarks are presented in Section 10.

## 2  Application Context and Approach

The notion of possibilistic expected utility may be applied in designing some decision making mechanisms in uncertain environments. One such an application is selection of most prospective partners in multi-agent negotiations [2] [3], where the notion of possibility based expected utility can be used to estimate the chance of a successful agreement with the potential negotiation partners. The main negotiator has to compose its service from components obtained from potential partners. She should select from the whole set of partners a subset of agents that can provide the best deals. To perform this task the main negotiator has to reason from historical negotiations about the negotiation capability of each potential partner. The approach requires construction of a possibility distribution for all potential partners through the process of case-based reasoning. The obtained possibility distribution is aggregated with the utility function and the resultant expected utilities for the modelled agents are then ordered. The order of these utilities determines the order of the agents from the most prospective to the least prospective one. However, the more attributes are taken into account the higher is the computational complexity of the algorithm constructing the distribution and deriving the expected utility. The complexity of the applied algorithm based on discretization is growing exponentially with the number of attributes (issues of negotiation) and with a high number of attributes the calculations become infeasible. Therefore an efficient algorithm for calculation of the expected utility is needed. The algorithm proposed in this paper is based on the observation that the shapes of $\alpha$-cuts of the distributions obtained by the possibilistic case-based reasoning are very regular, and by a common definition of the similarity relations (Section 3) they are unions of hypercuboids. This allows for estimating the expected utility without even constructing the possibility distribution. Two specific features help to derive such an efficient algorithm. First of them is the regular shape of $\alpha$-cuts and the second is the monotonicity of the utility function in the sense of Pareto. The remainder of this paper presents theoretical basis for the algorithm that efficiently calculates the possibility based expected utility.

## 3  Construction of similarity relation and case-based reasoning rule

In this section we present sample definition of similarity relation [8]. The theorems proved in next sections will use this kind of definition. The similarity of two values of $k$-th attribute can be defined as follows:

$$P_k(a,b) = f_k(|a-b|)$$

where function $f_k$ is some decreasing function mapping $R$ into $[0,1]$. The similarity relation comparing sequences of attributes: $\bar{o} = (o_1, o_2, \ldots, o_n)$ and $\bar{y} = (y_1, y_2, \ldots, y_n)$ is defined as an aggregation of atomic similarity relations:

$$\begin{aligned} P(\bar{o}, \bar{y}) &= P((o_1, o_2, \ldots, o_n), (y_1, y_2, \ldots, y_n)) = \\ & P_1(o_1, y_1) \otimes P_2(o_2, y_2) \otimes \ldots \otimes P_n(o_n, y_n) \end{aligned}$$

where $\otimes$ is a T-norm: $c \otimes d = Min[c,d]$. The density distribution function $\mu$ is obtained by the use of case-based reasoning rule stating that: "The more similar are the situation description attributes the more possible that the outcome attributes are similar":

$$\mu^t(\bar{y}) = Max_{i \in H^{t-1}} S(\bar{s}^i, \bar{s}^t) \otimes P(\bar{o}^i, \bar{y}) \qquad (1)$$

where S and P are similarity relations comparing situations and outcomes respectively, $H^{t-1}$ is the set of indexes of previous interactions, $\bar{s}^i = (s_1^i, s_2^i, \ldots, s_l^i)$ is the description of situation number $i$,
$\bar{s}^t = (s_1^t, s_2^t, \ldots, s_l^t)$ is the description of current situation and $\bar{o}^i = (o_1^i, o_2^i, \ldots, o_n^i)$ is the description of the outcome of the situation number $i$.

## 4 The $\alpha$-cuts of density distribution function

In this section we explore the $\alpha$-cuts of the function obtained by multi-dimensional case-based reasoning. We call this function density of possibility distribution and the function after transformation (presented in details in Section 6) will be called possibility distribution. This result will simplify the determination of the shapes of the $\alpha$-cuts of the function after transformation considered in the Section 7. It turns out that the $\alpha$-cuts of density distribution function obtained by possibilistic case-based reasoning are very regular and have shapes of union of rectangles in case of two-dimensional decision space. In a situation of multi-dimensional decision spaces the $\alpha$-cuts are unions of hipercuboids. We prove this property below.

**Theorem 1** *Let $\mu$ be $n$-dimensional denisty of possibility distribution obtained by possibilistic case-based reasoning:*

$$\begin{aligned} \mu(\bar{y}) &= \mu(y_1, y_2, \ldots, y_n) = \\ &= Max_{i \in H^{t-1}} S((s_1^i, s_2^i, \ldots, s_l^i), (s_1^t, s_2^t, \ldots, s_l^t)) \\ &\otimes P((o_1^i, o_2^i, \ldots, o_n^i), (y_1, y_2, \ldots, y_n)) \end{aligned}$$

*then the $\alpha$-cut*
$M_\alpha = \{(y_1, y_2, \ldots, y_n) \mid \mu(y_1, y_2, \ldots, y_n) \geq \alpha\}$ *is a union of hypercuboids:*

$$\begin{aligned} M_\alpha &= \bigcup_{i \in I^{t-1}} [o_1^i - f_1^{-1}(\alpha), o_1^i + f_1^{-1}(\alpha)] \times \ldots \\ &\times [o_n^i - f_n^{-1}(\alpha), o_n^i + f_n^{-1}(\alpha)] \end{aligned}$$

*where $I^{t-1}$ is a set of indexes of historical cases for which inequality is satisfied*
$S^i = S((s_1^i, s_2^i, \ldots, s_l^i), (s_1^t, s_2^t, \ldots, s_l^t)) \geq \alpha$ *and $f_j$ are the functions building similarity relation $P$.*

**Proof**

$$\bar{y} = (y_1, y_2, \ldots, y_n) \in M_\alpha \Leftrightarrow$$
$$\Leftrightarrow \mu(y_1, y_2, \ldots, y_n) \geq \alpha \Leftrightarrow$$
$$\Leftrightarrow Max_{i \in H^{t-1}} S((s_1^i, \ldots, s_l^i), (s_1^t, \ldots, s_l^t))$$
$$\otimes P((o_1^i, \ldots, o_n^i), (y_1, \ldots, y_n)) \geq \alpha \Leftrightarrow$$
$$\Leftrightarrow \bigvee_{i \in H^{t-1}} S^i \otimes P((o_1^i, \ldots, o_n^i), (y_1, \ldots, y_n)) \geq \alpha$$
$$\Leftrightarrow \bigvee_{i \in H^{t-1}} (S^i \geq \alpha \wedge P_1(o_1^i, y_1) \geq \alpha \wedge \ldots \wedge$$
$$\wedge P_n(o_n^i, y_n) \geq \alpha) \Leftrightarrow$$
$$\Leftrightarrow \bigvee_{i \in H^{t-1}} (S^i \geq \alpha \wedge f_1(|o_1^i - y_1|) \geq \alpha \wedge$$
$$\ldots \wedge f_n(|o_n^i - y_n|) \geq \alpha) \Leftrightarrow$$
$$\Leftrightarrow \bigvee_{i \in H^{t-1}} (S^i \geq \alpha \wedge |o_1^i - y_1| < f_1^{-1}(\alpha) \wedge$$
$$\ldots \wedge |o_n^i - y_n| < f_n^{-1}(\alpha)) \Leftrightarrow$$
$$\Leftrightarrow \bigvee_{i \in I^{t-1}} (y_1, \ldots y_n) \in$$
$$\in [o_1^i - f_1^{-1}(\alpha), o_1^i + f_1^{-1}(\alpha)] \times$$
$$\ldots \times [o_n^i - f_n^{-1}(\alpha), o_n^i + f_n^{-1}(\alpha)] \Leftrightarrow$$
$$\Leftrightarrow (y_1, \ldots, y_n) \in$$
$$\in \bigcup_{i \in I^{t-1}} [o_1^i - f_1^{-1}(\alpha), o_1^i + f_1^{-1}(\alpha)] \times$$
$$\ldots \times [o_n^i - f_n^{-1}(\alpha), o_n^i + f_n^{-1}(\alpha)]$$

The result of theorem 1 is ilustrated in Figure 1

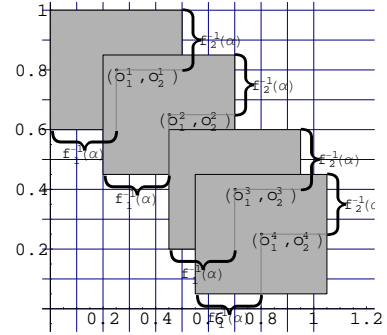

Figure 1: $\alpha$-cut $M_\alpha$ for two-dimensional space as a union of rectangles

## 5 Pareto order and monotonicity of multidimensional function

In this section we introduce some notions that will be required in the next sections. For the sake of simplicity we assume that all attributes have already been mapped into $[0, 1]$ intervals. In this situation the decision space is a cartesian product of all $n$ intervals: $D = [0, 1]^n$. $n$ is the number of all attributes for any agent that is modelled. From the point of view of the main negotiator that models its potential partners, some attributes can be positive (e.g. availability) or negative (e.g. price). That means that the utility function over domain of a positive attribute is increasing. Price is an example of the negative attribute because the more we pay the worse it is. Therefore the price utility function should be decreasing. We introduce now the notion of Pareto ordering. This tool will be applied later. Pareto ordering is a partial ordering which is useful for elimination of dominated alternatives. First we have to specify the point $\bar{b}$ in decision space which is best from a point of view of the main negotiator.

$$\bar{b} = [m_1, m_2, \ldots, m_n]$$

We assume that all considered attributes are monotone (eg. positive or negative). By such an assumption $m_j \in \{0,1\}$ for every $j \in \{1,2,\ldots,n\}$. For example, in a situation with two attributes (e.g. availability and price) we have $\bar{b} = \{1,0\}$. Two points $\bar{y} = (y_1, y_2, \ldots, y_n), \bar{z} = (z_1, z_2, \ldots, z_n)$ are in a weak Pareto relation

$$\bar{y} \preceq \bar{z} \Leftrightarrow \forall_{j \in \{1,2,\ldots,n\}} \quad (-1)^{m_j}(y_j - z_j) \geq 0$$

The weak Pareto order allows us to define the weak monotonicity of a multi-dimensional function. We say that the function $\pi$ of $n$ variables is weakly increasing (decreasing) if the following implication holds:

$$\bar{y} \preceq \bar{z} \Rightarrow \pi(\bar{y}) \leq \pi(\bar{z}) \quad (\bar{y} \preceq \bar{z} \Rightarrow \pi(\bar{y}) \geq \pi(\bar{z}))$$

We can also introduce the strong Pareto order. Two points $\bar{y} = (y_1, y_2, \ldots, y_n), \bar{z} = (z_1, z_2, \ldots, z_n)$ are in a strong Pareto relation

$$\bar{y} \prec \bar{z} \Leftrightarrow (\bar{y} \preceq \bar{z} \land \bar{y} \neq \bar{z})$$

and the function $\pi$ is increasing (decreasing):

$$\bar{y} \prec \bar{z} \Rightarrow \pi(\bar{y}) < \pi(\bar{z}) \quad (\bar{y} \prec \bar{z} \Rightarrow \pi(\bar{y}) > \pi(\bar{z}))$$

## 6 Transformation of distribution density function

In the case-based reasoning we obtain a function which does not satisfy the monotonicity condition. This function is supposed to estimate the utility function of a modelled agent which should be monotone. In other words, if we predict that an agent agrees on some value of positive attribute with some level of possibility $\alpha$ it should also agree on every smaller value with at least the same level of possibility $\alpha$. To obtain the nearest weak decreasing function $\pi$ greater or equal to $\mu$ we perform the following modification (Figure 2) known in the literature [6] as "upper anti-cumulative distribution":

$$\pi(\bar{y}) = sup_{\bar{y} \preceq \bar{x}} \mu(\bar{x})$$

**Theorem 2** $\pi(\bar{y}) = sup_{\bar{y} \preceq \bar{x}} \mu(\bar{x})$ *is the function nearest to $\mu$ satisfying condition:*

$$\forall \bar{y} \in D \,|\, \mu(\bar{y}) \leq \pi(\bar{y}) \land \bar{y} \preceq \bar{z} \Rightarrow \pi(\bar{y}) \geq \pi(\bar{z}) \quad (2)$$

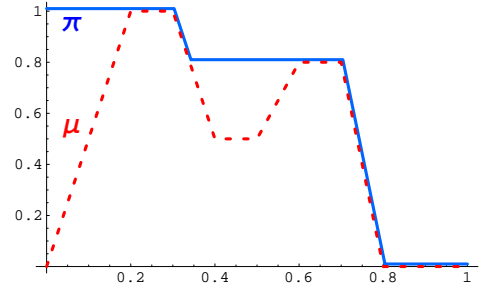

Figure 2: Illustration of transformation of density of possibility distribution to possibility distribution for one- dimensional decision space

**Proof** We choose any other function $\pi_1$ satisfying (2) and prove that it is greater or equal to $\pi$. ($\forall \bar{y} \in D \,|\, \pi(\bar{y}) \leq \pi_1(\bar{y})$)
Let $\bar{y} \in D$ and assume we found $\bar{x}_0$ ($\bar{y} \preceq \bar{x}_0$) for which $sup_{\bar{y} \preceq \bar{x}} \mu(\bar{x}) = \mu(\bar{x}_0)$ then

$$\pi_1(\bar{y}) - \pi(\bar{y}) = \pi_1(\bar{y}) - sup_{\bar{y} \preceq \bar{x}} \mu(\bar{x}) = \pi_1(\bar{y}) - \mu(\bar{x}_0)$$

but $\pi_1$ is monotonic ($\bar{y} \preceq \bar{x}_0 \Rightarrow \pi_1(\bar{y}) \geq \pi_1(\bar{x}_0)$) and greater or equal to $\mu$ therefore:

$$\pi_1(\bar{y}) - \mu(\bar{x}_0) \geq \pi_1(\bar{x}_0) - \mu(\bar{x}_0) \geq 0$$

what finaly allows us to conclude that $\forall \bar{y} \in D \,|\, \pi(\bar{y}) \leq \pi_1(\bar{y})$ and therefore $\pi$ is the nearest function to $\mu$ satysfying (2)

## 7 The $\alpha$-cuts of possibility distribution and Pareto frontier

In this section we analyse the shapes of $\alpha$-cuts of the possibility distribution obtained by transformation of the density distribution function, and define Pareto frontier. Let $M_\alpha$ and $\Pi_\alpha$ be the $\alpha$-cuts of the distribution density and distribution respectively:

$$M_\alpha = \{\bar{x} \quad | \quad \mu(\bar{x}) \geq \alpha\} \quad \Pi_\alpha = \{\bar{y} \quad | \quad \pi(\bar{y}) \geq \alpha\}$$

The following property holds:

**Theorem 3** $\bar{x} \in M_\alpha \Rightarrow \forall \bar{y} \quad | \quad \bar{y} \preceq \bar{x} \Rightarrow \bar{y} \in \Pi_\alpha$

**Proof** Let us choose any point $\bar{x}_0 \in M_\alpha$ and then let us choose any point $\bar{y}_0$ preccedeing or equivalent to $\bar{x}_0$: $\bar{y}_0 \preceq \bar{x}_0$, we have:

$$(\mu(\bar{x}_0) \geq \alpha \land \bar{y}_0 \preceq \bar{x}_0) \quad \Rightarrow \quad sup_{\bar{y}_0 \preceq \bar{x}} \mu(\bar{x}) \geq \alpha \Rightarrow$$
$$\pi(\bar{y}_0) \geq \alpha \Rightarrow \bar{y}_0 \in \Pi_\alpha$$

This theorem states that the distribution $\alpha$-cut $\Pi_\alpha$ contains the density $\alpha$-cut $M_\alpha$ and all points preceding the points belonging to $M_\alpha$.

**Theorem 4 (Conclusion)** *The distribution $\alpha$-cut $\Pi_\alpha$ is a union of hypercuboids:*

$$\Pi_\alpha = \bigcup_{i \in I^{t-1}} A_1^i(\alpha) \times \ldots \times A_n^i(\alpha)$$

*where $A_k^i(\alpha)$ is an interval of the form:*

$$A_k^i(\alpha) = \begin{cases} [0, o_k^i + f_k^{-1}(\alpha)] & for \quad m_k = 0 \\ [o_k^i - f_k^{-1}(\alpha), 1] & for \quad m_k = 1 \end{cases}$$

*as previously $I^{t-1}$ is a set of historical cases indexes for which the similarity degree $S^i$ exceeds $\alpha$ and $f_k$ are functions building the similarity relations.*

The Pareto frontier of the set $S$ is its subset denoted by $\mathcal{F}(S)$ containing points which are not dominated by other points from S.

$$\bar{x} \in \mathcal{F}(S) \iff \neg \exists \bar{y} \in S \mid \bar{x} \prec \bar{y}$$

In the case of our $\alpha$-cuts $\Pi_\alpha$ the Pareto frontier $F_\alpha = \mathcal{F}(\Pi_\alpha)$ is a set consisting of the vertices of the hypercuboids building the set $\Pi_\alpha$. Of course we mean the vertices nearest to the point $\bar{b} = [m_1, m_2, \ldots, m_n]$. This vertices are not dominated by other points of $\alpha$-cut and each point of $\Pi_\alpha$ is dominated by one of this vertices. Therefore, the following theorem is true:

**Theorem 5** *The Pareto frontier $F_\alpha$ of set $\Pi_\alpha$ consists of $|I^{t-1}|$ points and has a form:*

$$F_\alpha = \{(o_1^i + (-1)^{m_1} f_1^{-1}(\alpha), \ldots \\ \ldots, o_n^i + (-1)^{m_n} f_n^{-1}(\alpha)) | \quad i \in I^{t-1}\}$$

The theorems 4 and 5 are illustrated in Figure 3.

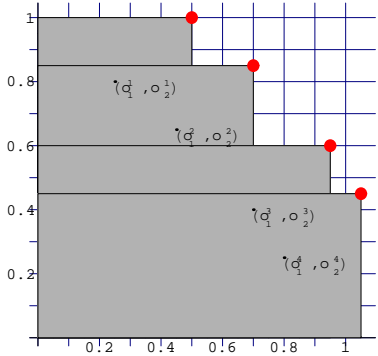

Figure 3: $\alpha$-cut $\Pi_\alpha$ for two-dimensional space and its Pareto frontier marked with circles

## 8 Qualitative expected utility

By the use of possibilistic case-based reasoning we obtain the possibility distribution $\pi$ modelling the partner (or partners) likelihood towards an agreement. To estimate the outcome of negotiation we have to aggregate it with the main negotiator's utility function $u$. By the aggregation we obtain a set $\mathcal{P}$ from which we find points $o \in O$ with the highest utility and treat it as prediction about possible negotiation outcomes (Figure 4). This type of aggregation is appropriate because it constitutes the confrontation of main negotiator's preferences encoded by his utility function with potential partner's capabilities encoded by the possibility distribution. The set $\mathcal{P}$ contains the points for which the utilities of both parties are predicted to be equal. The points in $O$ can be regarded as trade-off what means that we predict that the points in $\mathcal{P}$ are indifferent for the potential partner, therefore she can move towards agreements that are better for the main negotiator ($O$).

$$\mathcal{P} = argMax_{\bar{x} \in D} u(\bar{x}) \otimes \pi(\bar{x})$$
$$O = argMax_{\bar{y} \in \mathcal{P}} u(\bar{y})$$

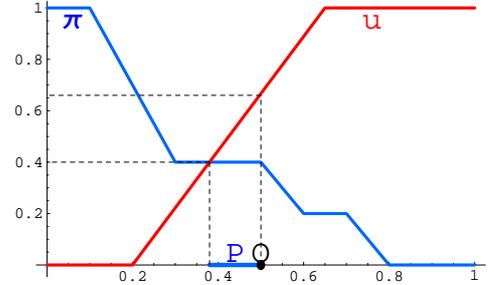

Figure 4: The ilustrating example showing the aggregation of distribution with utility function in one dimensional space. The set $\mathcal{P}$ is interval $\mathcal{P} = [0.4, 0.5]$ and set $\mathcal{O}$ consists of one point $\mathcal{O} = \{0.5\}$

The $\alpha$-cut $U_\alpha$ of the utility function $u$ may be treated as the main negotiator's zone of acceptable points with a level of satisfaction at least $\alpha$. Analogously the $\alpha$-cut $\Pi_\alpha$ of the possibility distribution may be treated as an estimation of partner's zone of acceptable points with a level of satisfaction at least $\alpha$. The intersection of these $\alpha$-cuts $U_\alpha \cap \Pi_\alpha$ may be considered as an estimation of zone of agreement on level $\alpha$ what is a set of points acceptable for both sides with level of satisfaction at least $\alpha$. The set $\mathcal{P}$ can be found in very efficient way by application of the notion of $\alpha$-cut. It turns out that the set $\mathcal{P}$ is equal to the zone of agreement on the highest possible level. In another words, to determine $\mathcal{P}$ we need to find the highest $\alpha$ for which the intersection of $\alpha$-cuts is not empty. The following theorem states it formally:

**Lemma 1** *Functions $u$ and $\pi$ are mapping the domain $D$ into interval $[0, 1]$, $U_\alpha$ and $\Pi_\alpha$ are its cor-*

responding $\alpha$-cuts $\Rightarrow \mathcal{P} = argMax_{\bar{x} \in D} u(\bar{x}) \otimes \pi(\bar{x}) = U_{\alpha_0} \cap \Pi_{\alpha_0} \wedge \alpha_0 = Max\{\alpha \,|\, U_\alpha \cap \Pi_\alpha \neq \emptyset\}$

The next lemma allows us to find the estimated negotiation outcome $o$ by maximization of the function $\mu$ over the $\alpha$-cut $\Pi_\alpha$ of the function $\pi$ instead of the set $\mathcal{P}$:

**Lemma 2** $U_{\alpha_0}$ and $\Pi_{\alpha_0}$ are as usually $\alpha_0$-cuts of functions $u$ and $\pi$ mapping the domain $D$ into interval $[0, 1]$. Following implication holds:

$$G_{\alpha_0} = U_{\alpha_0} \cap \Pi_{\alpha_0} \neq \emptyset \Rightarrow$$
$$argMax_{\bar{x} \in G_{\alpha_0}} u(\bar{x}) = argMax_{\bar{x} \in \Pi_{\alpha_0}} u(\bar{x})$$

The above lemma is illustrated in Figure 5

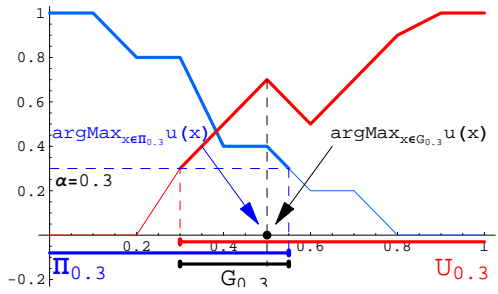

Figure 5: Illustration for lemma 2 in one-dimensional space $D = [0, 1]$

The next lemma states that the Pareto frontier $\mathcal{F}(S)$ of the set $S$ contains points for which the function $u$ reaches the highest value over $S$, if it is monotonic.

**Lemma 3** If $u : D \to [0, 1]$ is increasing ($\bar{x}_1 \prec \bar{x}_2 \Rightarrow u(\bar{x}_1) < u(\bar{x}_2)$), $S \subset D$ and $\mathcal{F}(S)$ is Pareto frontier of the set $S$ then the following inclusion holds:

$$argMax_{\bar{x} \in S} u(\bar{x}) \subset \mathcal{F}(S)$$

The above lemma is illustrated in Figure 6

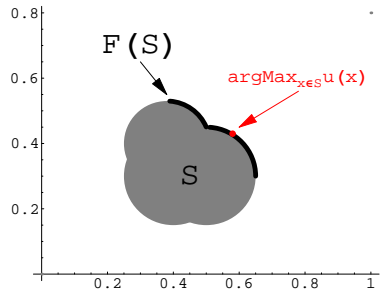

Figure 6: Ilustration for lemma 3 in two-dimensional space $D = [0, 1]^2$

**Lemma 4** If $U_\alpha$ is $\alpha$-cut of the function $u$ mapping the domain $D$ into an interval $[0, 1]$, $S \subset D$ and $u$ is increasing ($\bar{x}_1 \prec \bar{x}_2 \Rightarrow u(\bar{x}_1) < u(\bar{x}_2)$) then the following condition is satisified:

$$U_\alpha \cap S \neq \emptyset \Leftrightarrow U_\alpha \cap \mathcal{F}(S) \neq \emptyset$$

The above lemma is illustrated in Figure 7 The fol-lowing theorem is the core of the proposed algorithm calculating the estimation of the possible negotiation outcomes. It simplifies the determination of the set $O$ of the estimated outcomes.

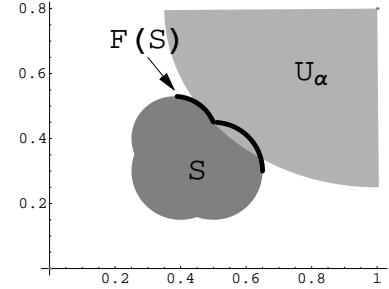

Figure 7: Illustration for lemma 4 in two-dimensional space $D = [0, 1]^2$

**Theorem 6** Assume that $u$ and $\pi$ are mapping domain $D$ into $[0, 1]$, the function $u$ is increasing ($\bar{x}_1 \prec \bar{x}_2 \Rightarrow u(\bar{x}_1) < u(\bar{x}_2)$), $F_{\alpha_0}$ is Pareto frontier of the functions $\pi$ $\alpha_0$-cut - $\Pi_{\alpha_0}$. The following statement holds:

$$\begin{aligned} If \quad & O = argMax_{\bar{y} \in \mathcal{P}} u(\bar{y}) \\ & where \quad \mathcal{P} = argMax_{\bar{x} \in D} u(\bar{x}) \otimes \pi(\bar{x}) \\ then \quad & O = argMax_{\bar{y} \in F_{\alpha_0}} u(\bar{y}) \\ & where \quad \alpha_0 = \{\alpha \,|\, Max_{\bar{x} \in F_\alpha} u(\bar{x}) \geq \alpha\} \end{aligned}$$

**Proof** Assume that $O = argMax_{\bar{y} \in \mathcal{P}} u(\bar{y})$ where $\mathcal{P} = argMax_{\bar{x} \in D} u(\bar{x}) \otimes \pi(\bar{x})$. Based on lemma (1) we can determine the set $\mathcal{P}$ using the notion of $\alpha$-cuts:

$$\mathcal{P} = U_{\alpha_0} \cap \Pi_{\alpha_0} \quad \wedge \quad \alpha_0 = Max\{\alpha \,|\, U_\alpha \cap \Pi_\alpha \neq \emptyset\}$$

then applying lemma (2) and (3) we obtain:

$$O = argMax_{\bar{y} \in \mathcal{P}} u(\bar{y}) = argMax_{\bar{y} \in \Pi_{\alpha_0}} u(\bar{y}) \subset F_{\alpha_0}$$

what gives us

$$\begin{aligned} O & = argMax_{\bar{y} \in F_{\alpha_0}} u(\bar{y}) \\ & where \quad \alpha_0 = Max\{\alpha \,|\, U_\alpha \cap \Pi_\alpha \neq \emptyset\} \end{aligned}$$

but based on lemma (4) we have:

$$U_\alpha \cap \Pi_\alpha \neq \emptyset \Leftrightarrow U_\alpha \cap F_\alpha \neq \emptyset \Leftrightarrow$$
$$\Leftrightarrow \exists \bar{x} \,|\, u(\bar{x}) \geq \alpha \wedge \bar{x} \in F_\alpha \Leftrightarrow Max_{\bar{x} \in F_\alpha} u(\bar{x}) \geq \alpha.$$

Therefore,

$$\begin{aligned} \alpha_0 & = Max\{\alpha \,|\, U_\alpha \cap \Pi_\alpha \neq \emptyset\} = \\ & = Max\{\alpha \,|\, Max_{\bar{x} \in F_\alpha} u(\bar{x}) \geq \alpha\} \end{aligned}$$

## 9 The algorithm

The theorem 6 allows us calculating the outcome $O$ in a very efficient way. We start from the highest $\alpha = 1$ and decrease it gradually, checking for a current degree, if a point for which $u$ exceeds the current level of $\alpha$ exists in the set $F_\alpha$. The set $F_\alpha$ is finite what allows us checking each point in it. If for some level $\alpha$ we find a point in $F_\alpha$ for which the function $u$ exceeds this level then this point is the solution. Assuming that $F_\alpha$ consists of $n$ points: $F_\alpha = \{f_\alpha^1, f_\alpha^2, \ldots, f_\alpha^n\}$ the algorithm can be summarised as follows:

```
STOP = FALSE;
α = 1;
Δα = 0.01;
while(!STOP)
{
   s = 0;
   for(i = 1; i <= n; i++)
   {
      if(u(f_α^i) > s)   s = u(f_α^i);
   }
   if(s >= α)   STOP = TRUE
   else   α = α - Δα;
}
```

This algorithm outperforms the classical approach [2][3]. The table 1 shows the comparison of computational complexity of the two approaches. It can be easily noticed that the complexity of classical approach grows exponentially with the number of attributes, whereas the complexity of the proposed algorithm grows just linearly.

Table 1: Computational complexity for both approaches. The timing is presented for the old and new approach.

| att | old | | | new |
|---|---|---|---|---|
| | distr | calcul | sum | estim |
| 2 | 2.72 | 0.12 | 2.84 | 0.06 |
| 3 | 41.88 | 1.65 | 43.53 | 0.09 |
| 4 | 539.93 | 20.95 | 560.88 | 0.11 |
| 5 | 6758.12 | 254.96 | 7013.08 | 0.14 |

## 10 Conclusions

In this paper we have proposed an efficient algorithm for estimation of possibility based qualitative expected utility. It explores the shapes of $\alpha$-cuts of the possibility distribution obtained through case-based reasoning. These results together with monotonicity of the utility function have been used for deriving a theoretical background of the proposed algorithm. The algorithm has been applied in the context of selecting the most prospective partners in multi-party multi-attribute negotiation, and can also be used in making decisions about potential offers during the negotiation based on a history of the previous interactions. Such an application requires construction of the possibility distribution for each potential partner, and aggregation of this distribution with the utility function. This aggregation is equivalent to the expected utility calculation and our algorithm allows for estimation of this utility in a very efficient way in comparison to other approaches.

## Acknowledgments

This work is part of the Adaptive Service Agreement and Process Management (ASAPM) in Services Grid project (AU-DEST-CG060081) in collaboration with the EU FP6 Integrated Project on Adaptive Services Grid (EU-IST-004617). The ASAPM project is proudly supported by the *Innovation Access Programme - International Science and Technology* established under the Australian Governments innovation statement, *Backing Australia's Ability*.